\documentclass[11pt]{article}

\usepackage{EMNLP2022}

\usepackage{times}
\usepackage{latexsym}

\usepackage[T1]{fontenc}

\usepackage[utf8]{inputenc}

\usepackage{microtype}

\usepackage{subfigure}
\usepackage{amsmath}
\usepackage{multirow}
\usepackage[subpreambles=true]{standalone}
\usepackage{import}
\usepackage{graphicx}
\usepackage{makecell}

\newcommand\blfootnote[1]{%
\begingroup
\renewcommand\thefootnote{}\footnote{#1}%
\addtocounter{footnote}{-1}%
\endgroup
}

\newcommand{\hytt}[1]{\texttt{\hyphenchar \font=\defaulthyphenchar#1}}

\title{PromptBERT: Improving BERT Sentence Embeddings with Prompts}
 \author{
     Ting Jiang\textsuperscript{\rm 1,$*$}, Jian Jiao\textsuperscript{\rm 3}, Shaohan Huang\textsuperscript{\rm 3}, Zihan Zhang\textsuperscript{\rm 3}, Deqing Wang\textsuperscript{\rm 1,4,}$^\dagger$,\\ {\bf Fuzhen Zhuang\textsuperscript{\rm 1,2,4} }, {\bf Furu Wei\textsuperscript{\rm 3}}, {\bf Haizhen Huang\textsuperscript{\rm 3}}, {\bf Denvy Deng\textsuperscript{\rm 3} }, {\bf Qi Zhang\textsuperscript{\rm 3} }\\
     \textsuperscript{\rm 1}SKLSDE Lab, School of Computer, Beihang University, Beijing, China \\
     \textsuperscript{\rm 2}Institute of Artificial Intelligence, Beihang University, Beijing, China\\
     \textsuperscript{\rm 3} Microsoft \textsuperscript{\rm 4} Zhongguancun Laboratory, Beijing, China \\
     \{royokong, dqwang, zhuangfuzhen\}@buaa.edu.cn\\
     \{shaohanh, zihzha, fuwei, hhuang, liazha, zhang.qi\}@microsoft.com
 }

\begin{document}
\maketitle
\begin{abstract}
 \blfootnote{* Work done during internship at microsoft.}
 \blfootnote{ $\dagger$ Corresponding Author.}

We propose PromptBERT, a novel contrastive learning method for learning better sentence representation. We firstly analyze the drawback of current sentence embedding from original BERT and find that it is mainly due to the static token embedding bias and ineffective BERT layers.
Then we propose the first prompt-based sentence embeddings method and discuss two prompt representing methods and three prompt searching methods to make BERT achieve better sentence embeddings.
Moreover, we propose a novel unsupervised training objective by the technology of template denoising, which substantially shortens the performance gap between the supervised and unsupervised settings. Extensive experiments show the effectiveness of our method. Compared to SimCSE, PromptBert achieves 2.29 and 2.58 points of improvement based on BERT and RoBERTa in the unsupervised setting. Our code is available at \url{https://github.com/kongds/Prompt-BERT}.
\end{abstract}

\section{Introduction}
In recent years, we have witnessed the success of pre-trained language models like BERT~\cite{devlin2018bert} and RoBERTa~\cite{liu2019roberta} in sentence embeddings~\cite{gao2021simcse, yan2021consert}.
However,  original BERT still shows poor performance in sentence embeddings \cite{reimers2019sentence, li2020sentence}. The most commonly used example is that it underperforms the traditional word embedding methods like GloVe \cite{pennington2014glove}.

Previous research has linked anisotropy to explain the poor performance of original BERT~\cite{li2020sentence, yan2021consert, gao2021simcse}.
Anisotropy makes the token embeddings occupy a narrow cone, resulting in a high similarity between any sentence pair~\cite{li2020sentence}. \citeauthor{li2020sentence}~(\citeyear{li2020sentence}) proposed a normalizing flows method to transform the sentence embeddings distribution to a smooth and isotropic Gaussian distribution and \citeauthor{yan2021consert}~(\citeyear{yan2021consert}) presented a contrastive framework to transfer sentence representation. The goal of these methods is to eliminate anisotropy in sentence embeddings.
However, we find that anisotropy may not be the primary cause of poor semantic similarity. For example, averaging the last layer of original BERT is even worse than averaging its static token embeddings in semantic textual similarity task, but the sentence embeddings from last layer are less anisotropic than static token embeddings.

Following this result, we find original BERT layers actually damage the quality of sentence embeddings. However, if we treat static token embeddings
as word embedding, it still yields unsatisfactory results compared to GloVe.
Inspired by~\cite{li2020sentence}, who found token frequency biases its distribution, we find the distribution of token embeddings is not only biased by frequency, but also case sensitive and subword in WordPiece \cite{wu2016google}. We design a simple experiment to test our conjecture by simply removing these biased tokens (e.g., high frequency subwords and punctuation) and using the average of the remaining token embeddings as sentence representation.
It can outperform the Glove and even achieve results comparable to post-processing methods BERT-flow~\cite{li2020sentence}  and BERT-whitening~\cite{su2021whitening}.

Motivated by these findings, avoiding embedding bias can improve the performance of sentence representations. However, it is labor-intensive to manually remove embedding biases and it may result in the omission of some meaningful words if the sentence is too short. Inspired by \cite{brown2020language}, which has reformulated the different NLP tasks as fill-in-the-blanks problems by different prompts,
we propose a prompt-based method by using the template to obtain the sentence representations in BERT. Prompt-based method can avoid embedding bias and utilize the original BERT layers.
We find original BERT can achieve reasonable performance with the help of the template in sentence embeddings, and it even outperforms some BERT based methods, which fine-tune BERT in down-stream tasks.

Our approach is equally applicable to fine-tuned settings.
Current methods utilize contrastive learning to help the BERT learn better sentence embeddings~\cite{gao2021simcse,yan2021consert}. However, the unsupervised methods still suffer from leaking proper positive pairs. \citeauthor{yan2021consert}~(\citeyear{yan2021consert}) discuss four data augmentation methods, but the performance seems worse than directly using the dropout in BERT as noise \cite{gao2021simcse}.
We find prompts can provide a better way to generate positive pairs by different viewpoints from different templates. To this end, we propose a prompt based contrastive learning method with template denoising to leverage the power of BERT in an unsupervised setting, which significantly shortens the gap between the supervised and unsupervised performance. Our method achieves state-of-the-art results in both unsupervised and supervised settings.

 \section{Related Work}
 Learning sentence embeddings as a fundamental NLP problem has been largely studied. Currently, how to leverage the power of BERT in sentence embeddings has become a new trend. Many works~\cite{li2020sentence, gao2021simcse} achieved strong performance with BERT in both supervised and unsupervised settings.
 Among these works, contrastive learning based methods achieve state-of-the-art results.
 \citeauthor{gao2021simcse}~(\citeyear{gao2021simcse}) proposed a novel contrastive training objective to directly use inner dropout as noise to construct positive pairs. \citeauthor{yan2021consert}~(\citeyear{yan2021consert}) discussed four methods to construct positive pairs.

 Although BERT achieved success in sentence embeddings, original BERT shows unsatisfactory performance~\cite{reimers2019sentence, li2020sentence}.
 One explanation is the anisotropy in original BERT, which causes sentence pairs to have high similarity, some works~\cite{li2020sentence, su2021whitening} focused on reducing the anisotropy by post-processing sentence embeddings.

\section{Rethinking the Sentence Embeddings of the Original BERT}\label{sec:finding}
Previous works~\cite{yan2021consert, gao2021simcse} explained the poor performance of the original BERT is mainly due to the learned anisotropic token embeddings space, where the token embeddings occupy a narrow cone.
However, we find that anisotropy is not a key factor to inducing poor semantic similarity by examining the relationship between the aniostropy and performance. We think the main reasons are the ineffective BERT layers and static token embedding biases.

\textbf{Observation 1: Original BERT layers fail to improve the performance.}
In this section, we analyze the influence of BERT layers by comparing the two sentence embedding methods: averaging static token embeddings (input of the BERT layers) and averaging last layer (output of the BERT layers). We report the sentence embedding performance and its sentence level anisotropy.

To measure the anisotropy, we follow the work of~\citeauthor{ethayarajh2019contextual}~(\citeyear{ethayarajh2019contextual}) to measure the sentence level anisotropy in sentence embeddings. Let \(s_i\) be a sentence that appears in corpus \(\{s_1,...,s_n\}\). The anisotropy can be measured as follows:
\begin{equation}\label{eq:anisotropy}
\frac{1}{n^{2}-n} \left| \sum_{i} \sum_{j \neq i} \cos \left(M(s_i),M(s_j)\right) \right|
\end{equation}
where \(M\) denotes the sentence embedding method, which maps the raw sentence to its embedding and \(cos\) is the cosine similarity.   In other words, the anisotropy of \(M\) is measured by the average cosine similarity of a set of sentences.
If sentence embeddings are isotropic (i.e., directionally uniform), then the average cosine similarity between uniformly randomly sampled sentences would be 0~\cite{arora2016simple}. The closer it is to 1, the more anisotropic the embedding of sentences.
We randomly sample 100,000 sentences from the Wikipedia corpus to compute the anisotropy.


We compare different pre-trained models (\hytt{bert-base-uncased}, \hytt{bert-base-cased} and \hytt{roberta-base}) in combination with different sentence embedding methods ( last layer average, averaging of last hidden layer tokens as sentence embeddings and static token embeddings, directly averaging of static token embeddings).
We list the spearman correlation and sentence level anisotropy of each combination in Table \ref{tab:anisotropy_comp}.

\begin{table}[h]
\centering
\begin{tabular}{lcc}
\hline
\multirow{2}*{Pre-trained models}                & Spearman & Sentence \\
& correlation & anisotropy \\
\hline
 \multicolumn{3}{c}{\textit{Static token embeddings avg.}}  \\
\small{\hytt{bert-base-uncased}} & 56.02                      & 0.8250 \\
\small{\hytt{bert-base-cased}}   & 56.65                      & 0.5755 \\
\small{\hytt{roberta-base}}      & 55.88                      & 0.5693 \\
 \multicolumn{3}{c}{\textit{Last layer avg.}}  \\
\small{\hytt{bert-base-uncased}} & 52.57                      & 0.4874  \\
\small{\hytt{bert-base-cased}}   & 56.93                      & 0.7514  \\
\small{\hytt{roberta-base}}      & 53.49                      & 0.9554  \\
\hline
\end{tabular}
\caption{ Spearman correlation and sentence anisotropy from static token embeddings averaging and last layer averaging. The spearman correlation is the average of STS12-16, STS-B and SICK.
} \label{tab:anisotropy_comp}
\end{table}

Table \ref{tab:anisotropy_comp} shows the BERT layers in \hytt{bert-base-uncased} and \hytt{roberta-base} significantly harm the sentence embeddings performance. Even in \hytt{bert-base-cased}, the gain of BERT layers is trivial with only 0.28 improvement.
We also show the sentence level anisotropy of each method. The performance degradation of the BERT layers seems not to be related to the sentence level anisotropy.
For example, the last layer averaging is more isotropic than the static token embeddings averaging in \hytt{bert-base-uncased}. However, the static token embeddings average achieves better sentence embeddings performance.



\begin{figure*}[h]
\centering    

\subfigure[Frequency bias in \hytt{bert-base-uncased}.]
{
	\includegraphics[width=0.64\columnwidth]{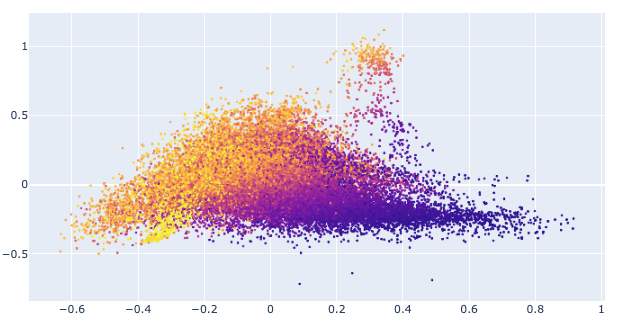}
}
\subfigure[Frequency bias in \hytt{bert-base-cased}.]
{
	\includegraphics[width=0.64\columnwidth]{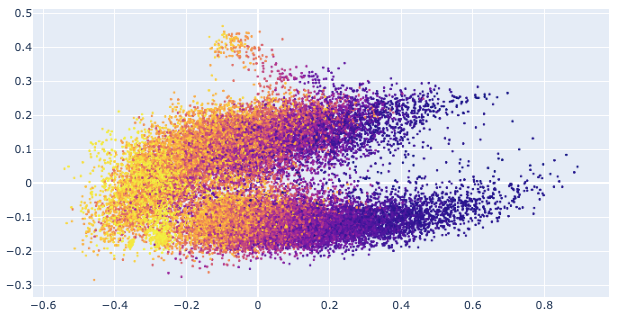}
}
\subfigure[Frequency bias in \hytt{roberta-base}.] 
{
	\includegraphics[width=0.64\columnwidth]{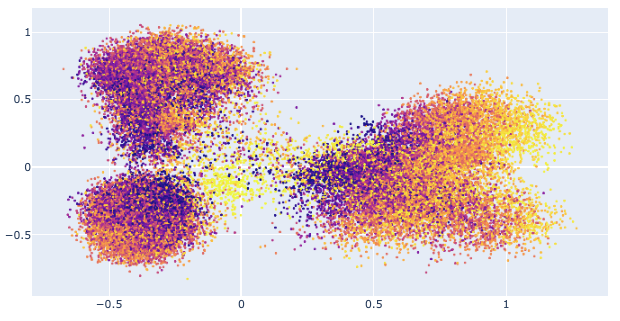}   
}

\subfigure[Subword and Case biases in \hytt{bert-base-uncased}.]
{
	\includegraphics[width=0.64\columnwidth]{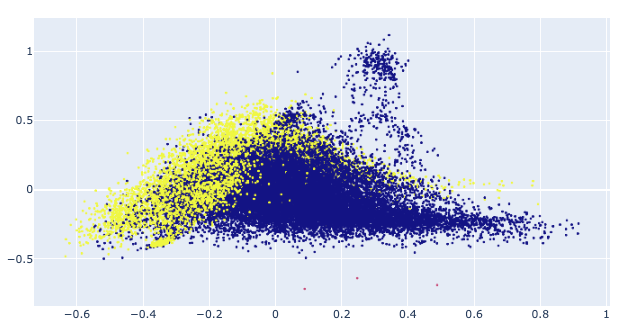}
}
\subfigure[Subword and Case biases in \hytt{bert-base-cased}.]
{
	\includegraphics[width=0.64\columnwidth]{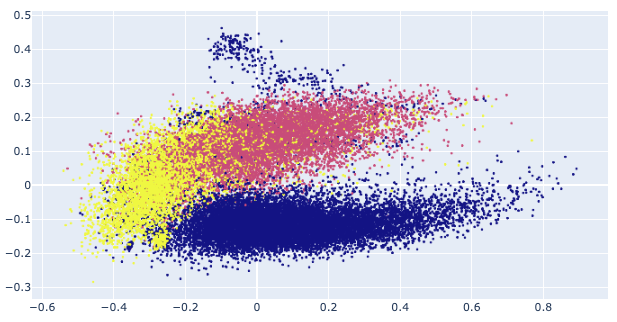}
}
\subfigure[Subword and Case biases in \hytt{roberta-base}.] 
{
	\includegraphics[width=0.64\columnwidth]{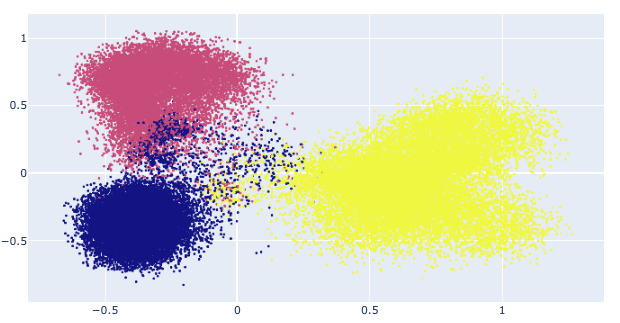}   
}
\hspace{0.66\columnwidth}

\caption{2D visualization of token embeddings with different biases. For frequency bias, the darker the color, the higher the token frequency. For subword and case bias, yellow represents subword and red represents the token contains capital letters.
} 
\label{fig:2dv}
\end{figure*}

\textbf{Observation 2: Embedding biases harms the sentence embeddings performance.} \label{sec:ob2}
\citeauthor{li2020sentence}~(\citeyear{li2020sentence}) found that token embeddings can be biased to token frequency. Similar problems have been studied in~\cite{yan2021consert}. The anisotropy in BERT static token embeddings is sensitive to token frequency. Therefore, we investigate whether embedding bias yields unsatisfactory performance of sentence embeddings. We observe that the token embeddings is not only biased by token frequency, but also subwords in WordPiece~\cite{wu2016google} and case sensitive. 

As shown in Figure~\ref{fig:2dv}, we visualize these biases in the token embeddings of \hytt{bert-base-uncased}, \hytt{bert-base-cased} and \hytt{roberta-base}. The token embeddings of three pre-trained models are highly biased by the token frequency, subword and case.
The token embeddings can be roughly divided into three regions according to the subword and case biases: 1) the lowercase begin-of-word tokens, 2) the uppercase begin-of-word tokens and 3) the subword tokens.
For uncased pre-trained model \hytt{bert-base-uncased}, the token embeddings also can be roughly divided into two regions: 1) the begin-of-word tokens, 2) the subword tokens.

For frequency bias, we can observe that high frequency tokens are clustered, while low frequency tokens are dispersed sparsely in all three models \cite{yan2021consert}. The begin-of-word tokens are more
vulnerable to frequency than subword tokens in BERT. However, the subword tokens are more vulnerable in RoBERTa.

Previous works \cite{yan2021consert, li2020sentence} often link token embeddings bias to the token embedding anisotropy and argue it is the main reason for the bias. However, we believe the anisotropy is unrelated to the bias.
The bias means the distribution of embedding is disturbed by some irrelevant information like token frequency, which can be directly visualized according to the PCA. For the anisotropy, it means the whole embedding occupies a narrow cone in the high dimensional vector space, which cannot be directly visualized.

\begin{table}[h]
\centering
\begin{tabular}{l|c}
\hline
\thead{$M$}                   & average cosine similarity\\
\hline
\small{\hytt{bert-base-uncased}} & 0.4445 \\
\small{\hytt{bert-base-cased}}   & 0.1465 \\
\small{\hytt{roberta-base}}       & 0.0235\\
\hline
\end{tabular}
\caption{ The average cosine similarity in static token embeddings
} \label{tab:acs}
\end{table}

Table \ref{tab:acs} shows the static token embeddings anisotropy of three pre-trained models in Figure \ref{fig:2dv} according  to the average the cosine similarity between any two token embeddings.
Contrary to the previous conclusion \cite{yan2021consert, li2020sentence}, we find only \hytt{bert-base-uncased}'s  static token embeddings is highly anisotropic. The static token embeddings like \hytt{roberta-base} are isotropic with 0.0235 average cosine similarity.
For biases, these models suffer from the biases in static token embeddings,  which is irrelevant to the anisotropy.

\begin{table}[h]
\centering
\setlength{\tabcolsep}{2pt}
\small
\begin{tabular}{l|ccc}
\hline & cased  & uncased & roberta\\
\hline
Static Token Embeddings  & 56.93 & 56.02 & 55.88 \\
~~~~$-$ Freq.   &  60.27 & 59.65 & 65.41\\
~~~~$-$ Freq. \& Sub.   & 64.83 & 62.20 & 64.89 \\
~~~~$-$ Freq. \& Sub. \& Case  & 65.07 & - & 65.06  \\
~~~~$-$ Freq. \& Sub. \& Case \& Pun.  & 66.05 & 63.10 & 67.64 \\
\hline
\end{tabular}
\caption{The influence of static embedding biases in spearman correlation. The spearman correlation is the average of STS12-16, STS-B and SICK. Cased, uncased and roberta represent \hytt{bert-base-cased}, \hytt{bert-base-uncased} and \hytt{roberta-base}.
For Freq., Sub., Case. and Pun., we remove the top frequency tokens, subword tokens, uppercase tokens and punctuation respectively. More details can be found in Appendix~\ref{sec:appendix_sta_token}.
} \label{tab:bias}
\end{table}

To prove the negative impact of biases, we show the influence of biases to the sentence embeddings with averaging static token embeddings as sentence embeddings (without BERT layers).
The results of eliminating embedding biases are quite impressive on three pre-trained models in Table \ref{tab:bias}. Simply removing a set of tokens, the result can be improved by 9.22, 7.08 and 11.76 respectively. The final result of \hytt{roberta-base} can outperform post-processing methods such as BERT-flow \cite{li2020sentence} and BERT-whitening \cite{su2021whitening} with only using static token embeddings.

Manually removing embedding biases is a simple method to improve the performance of sentence embeddings. However, if the sentence is too short, this is not an adequate solution, which may result in the omission of some meaningful words.

\section{Prompt Based Sentence Embeddings}
Inspired by \citeauthor{brown2020language}~(\citeyear{brown2020language}), we propose a prompt based sentence method to obtain sentence embeddings.
By reformulating the sentence embedding task as the mask language task, we can effectively use original BERT layers by leveraging the large-scale knowledge. We also avoid the embedding biases by representing sentences from [MASK] tokens.

However, unlike the text classification and question-answering tasks, the output in sentence embeddings is not the label tokens predicted by MLM classification head, but the vector to represent the sentence. We discuss the implementation of prompt based sentence embeddings through addressing the following two questions:
1) how to represent sentences with the prompt, and
2) how to find a proper prompt for sentence embeddings.
Based on these, we propose a prompt based contrastive learning method to fine-tuning BERT on sentence embeddings.
\subsection{Represent Sentence with the Prompt}
In this section, we discuss two methods to represent one sentence with a prompt.
For example, we have a template ``[X] means [MASK]'', where [X] is a placeholder to put sentences and [MASK] represents the [MASK] token.
Given a sentence \(x_{\rm in}\), we map \(x_{\rm in}\) to \(x_{\rm prompt}\) with the template. Then we feed \(x_{\rm prompt}\) to a pre-trained model to generate sentence representation $\mathbf{h}$.

One method is to use the hidden vector of [MASK] token as sentence representation:
\begin{equation}\label{eq:anisotropy1}
\mathbf{h} = \mathbf{h}_{[{\rm MASK}]}
\end{equation}

For the second method like other prompt based tasks, we get the top-\(k\) tokens according to \(\mathbf{h}_{[{\rm MASK}]}\) and MLM classification head, then calculate the weighted average of these tokens according to probability distribution. The \(\mathbf{h}\) can be formulated as:
\begin{equation}\label{eq:anisotropy2}
\mathbf{h} = \frac{\sum_{v \in \mathcal{V}_{\text{top-}k}} \mathbf{W}_v \operatorname{P}\left(v = {\rm [MASK]} | \mathbf{h}_{\rm [MASK]}\right)}{\sum_{v \in \mathcal{V}_{\text{top-}k}} \operatorname{P}\left(v = {\rm [MASK]} | \mathbf{h}_{\rm [MASK]}\right)}
\end{equation}
where \(v\) is the BERT token in the \(\text{top-}k\) tokens set \(\mathcal{V}_{\text{top-}k}\), \(\mathbf{W}_v\) is the static token embeddings of \(v\) and \(\operatorname{P}\left(v = {\rm [MASK]} | \mathbf{h}_{\rm [MASK]}\right)\) denotes the probability of token \(v\) be predicted by MLM head with \(\mathbf{h}_{\rm [MASK]}\).

The second method, which maps the sentence to the tokens, is more conventional than the first. But its disadvantages are obvious: 1) as previously noted, due to the sentence embeddings from averaging of static token embeddings, it still suffers from biases. 2) weight averaging makes the BERT hard to fine-tune in down-stream tasks. For these reasons, we represent the sentence with the prompt by the first method.

\subsection{Prompt Search}
For prompt based tasks, one key challenge is to find templates.
We discuss three methods to search for templates in this section: manual search, template generation based on T5~\cite{gao2020making} and OptiPrompt~\cite{zhong2021factual}.
We use the spearman correlation in the STS-B development set as the main metric to evaluate different templates.

For manual search, we need to hand-craft templates and give a strong hint that the whole sentence is represented as \(\mathbf{h}_{\rm [MASK]}\).
To search templates, we divide the template into two parts: relationship tokens, which denote the relationship between [X] and [MASK], and prefix tokens, which wrap [X]. Then we greedily search for templates following the relationship tokens and prefix tokens.

\begin{table}[h]
\centering
\small
\begin{tabular}{lc}
\hline
Template                                 & STS-B dev.  \\
\hline
\multicolumn{2}{c}{\textit{Searching for relationship tokens }} \\
{}[X] [MASK] .                           & 39.34 \\
{}[X] is [MASK] .                        & 47.26 \\
{}[X] mean [MASK] .                      & 53.94 \\
{}[X] means [MASK] .                     & 63.56 \\
\hline
\multicolumn{2}{c}{\textit{Searching for prefix tokens }} \\
This [X] means [MASK] .                  & 64.19\\
This sentence of [X] means [MASK] .      & 68.97\\
This sentence of ``[X]'' means [MASK] . & 70.19\\
This sentence : ``[X]'' means [MASK] .  & 73.44\\
\hline
\end{tabular}
\caption{ Greedy searching templates on \hytt{bert-base-uncased}.
} \label{tab:gss}
\end{table}

Some results of greedy searching are shown in Table~\ref{tab:gss}.
When it comes to sentence embeddings, different templates produce extremely varied results. Compared to simply concatenating the [X] and [MASK], complex templates like \textit{This sentence : ``}[X]\textit{'' means }[MASK]., can improve the spearman correlation by 34.10.

For template generation based on T5, \citeauthor{gao2020making}~(\citeyear{gao2020making}) proposed a novel method to automatically generate templates by using T5 to generate templates according to the sentences and corresponding labels. The generated templates can outperform the manual searched templates in the GLUE benchmark \cite{wang2018glue}.

However, the main issue to implement it is the lack of label tokens. \citeauthor{tsukagoshi2021defsent}~(\citeyear{tsukagoshi2021defsent}) successfully transformed the sentence embeddings task to the text classification task by classifying the definition sentence to its word according to the dictionary. Inspired by this, we use words and corresponding definitions to generate 500 templates (e.g., orange: a large round juicy citrus fruit with a tough bright reddish-yellow rind).
Then we evaluate these templates in the STS-B development set, the best spearman correlation is 64.75 with the template ``Also called [MASK]. [X]''.
Perhaps it is the gap between sentence embeddings and word definition. This method cannot generate better templates compared to manual searching.

OptiPrompt \cite{zhong2021factual} replaced discrete template with the continuous template. To optimize the continuous template, we use the unsupervised contrastive learning as training objective following the settings in \citeauthor{gao2021simcse}~(\citeyear{gao2021simcse}) with freezing the whole BERT parameters, and the continuous template is initialized by manual template's static token embeddings. Compared to the input manual template, the continuous template can increase the spearman correlation from 73.44 to 80.90 on STS-B development set.

\subsection{Prompt Based Contrastive Learning with Template Denoising}
Recently, contrastive learning successfully leverages the power of BERT in sentence embeddings. A challenge in sentence embeddings contrastive learning is how to construct proper positive instances. \citeauthor{gao2021simcse}~(\citeyear{gao2021simcse}) directly used the dropout in the BERT as positive instances. \citeauthor{yan2021consert}~(\citeyear{yan2021consert}) discussed the four data augmentation strategies such as adversarial attack, token shuffling, cutoff and dropout in the input token embeddings to construct positive instances.
Motivated by the prompt based sentence embeddings, we propose a novel method to reasonably generate positive instances based on prompt.

The idea is using the different templates to represent the same sentence as different points of view, which helps model to produce more reasonable positive pairs.
To reduce the influence of the template itself on the sentence representation, we propose a novel way to denoise the template information. Given the sentence \(x_{i}\), we first calculate the corresponding sentence embeddings \(\mathbf{h}_{i}\) with a template. Then we calculate the template bias \(\hat{\mathbf{h}}_{i}\) by directly feeding BERT with the template and the same template position ids. For example, if the \(x_{i}\) has 5 tokens, then the position ids of template tokens after the [X] will be added by 5 to make sure the position ids of template are same. Finally, we can directly use the \(\mathbf{h}_{i} - \hat{\mathbf{h}}_{i}\) as the denoised sentence representation. For the template denoising, more details can be found in Discussion.

Formally,
let \(\mathbf{h}_{i}^{'}\) and \(\mathbf{h}_{i}\) denote the sentence embeddings of \(x_{i}\) with different templates, \(\hat{\mathbf{h}}_{i}^{'}\) and \(\hat{\mathbf{h}}_{i}\) denotes the two template biases of the \(x_{i}\) respectively, the final training objective is as follows:

\begin{equation}\label{eq:trainobj}
\ell_{i}=-\log \frac{e^{\operatorname{cos}\left(\mathbf{h}_{i} - \hat{\mathbf{h}}_{i}, \mathbf{h}_{i}^{'} - \hat{\mathbf{h}}_{i}^{'}\right) / \tau}}{\sum_{j=1}^{N} e^{\operatorname{cos}\left(\mathbf{h}_{i} - \hat{\mathbf{h}}_{i}, \mathbf{h}_{j}^{'} - \hat{\mathbf{h}}_{j}^{'}\right) / \tau}}
\end{equation}
where \(\tau\) is a temperature hyperparameter in contrastive learning and \(N\) is the size of mini-batch.

\section{Experiments}
We conduct experiments on STS tasks with non fine-tuned and fine-tuned BERT settings. For non fine-tuned BERT settings, we exploit the performance of original BERT in sentence embeddings, which corresponds to the previous findings of the poor performance of original BERT.
For fine-tuned BERT settings, we report the unsupervised and supervised results by fine-tuning BERT with downstream tasks. The results of transfer tasks are in Appendix \ref{sec:appendix_tt}.

\begin{table*}[h!]
\renewcommand\arraystretch{1.1}
\centering
\small
\setlength{\tabcolsep}{5pt}
\begin{tabular}{lcccccccc}
\hline
Method & STS12 & STS13 & STS14 & STS15 & STS16 & STS-B & SICK-R & Avg.\\
\hline
GLoVe embeddings avg.\(^\dagger\) & 55.14 & 70.66 & 59.73 & 68.25 & 63.66 & 58.02 & 53.76 & 61.32  \\
BERT last avg. & 30.87 & 59.89 & 47.73 & 60.29 & 63.73 & 47.29 & 58.22 & 52.57 \\ 
\hspace{2.5em} static avg. & 42.38 & 56.74 & 50.60 & 65.08 & 62.39 & 56.82 & 58.15 & 56.02 \\
\hspace{2.5em} first-last avg.\(^\dagger\) & 39.70 & 59.38 & 49.67 & 66.03 & 66.19 & 53.87 & 62.06 & 56.70\\
\hspace{2.5em} static remove biases avg. & 53.09 & 66.48 & 65.09 & 69.80 & 67.85 & 61.60 & 57.80 & 63.10 \\
BERT-flow \(^\dagger\) & 58.40 & 67.10 & 60.85 & 75.16 & 71.22 & 68.66 & 64.47 & 66.55 \\
BERT-whitening  \(^\dagger\) & 57.83 & 66.90 & 60.90 & 75.08 & 71.31 & 68.24 & 63.73 & 66.28 \\
Prompt based BERT (manual) & 60.96 & 73.83 & 62.18 & 71.54 & 68.68 & 70.60 & 67.16 & 67.85 \\
Prompt based BERT (manual\&OptiPrompt)&  \textbf{64.56} & \textbf{79.96} & \textbf{70.05} & \textbf{79.37} & \textbf{75.35} & \textbf{77.25} & \textbf{68.56} & \textbf{73.59} \\
\hline
\end{tabular}
\caption{ The performance comparison of our unfine-tuned BERT method on STS tasks. \(^\dagger\): results from \cite{gao2021simcse}. The BERT-flow\cite{li2020sentence} and BERT-whitening \cite{su2021whitening} use the "NLI" setting. All BERT based methods use \hytt{bert-base-uncased} .  Last avg. denotes averaging the last layer of BERT. Static avg. denotes averaging the static token embedding of BERT.  First-last avg. \cite{su2021whitening} uses the first and last layer. Static remove biases avg. means removing biased tokens in static avg., which we have introduced before.
} \label{tab:ufr}
\end{table*}

\begin{table*}[!h]
\centering
\renewcommand\arraystretch{1.1}
\small
\setlength{\tabcolsep}{2pt}

\begin{tabular}{lcccccccc}
\hline
Method & STS12 & STS13 & STS14 & STS15 & STS16 & STS-B & SICK-R & Avg.\\
\hline
\hline
\multicolumn{9}{c}{\textit{Unsupervised models}} \\
\hline
IS-BERT$_{\textrm{base}}$$^\P$ & 56.77 & 69.24 & 61.21 & 75.23 & 70.16 & 69.21 & 64.25 & 66.58 \\
ConSERT$_{\textrm{base}}$\(^\ddagger \)  & 64.64 & 78.49 & 69.07 & 79.72 & 75.95 & 73.97 & 67.31 & 72.74 \\
SimCSE-BERT$_{\textrm{base}}$\(^\ddagger \)  & 68.40 & 82.41 & 74.38 & 80.91 & 78.56 &    76.85     &      \textbf{72.23}      & 76.25 \\
PromptBERT$_{\textrm{base}}$ & \textbf{71.56$_{\pm 0.18}$} & \textbf{84.58$_{\pm 0.22}$} & \textbf{76.98$_{\pm 0.26}$} & \textbf{84.47$_{\pm 0.24 }$} & \textbf{80.60$_{\pm 0.21}$} &  \textbf{81.60$_{\pm 0.22}$} & 69.87$_{\pm 0.40}$ & \textbf{78.54$_{\pm 0.15}$} \\
\hline
RoBERTa$_{\textrm{base}}$-whitening\(^\dagger\) & 46.99 & 63.24 & 57.23 & 71.36 & 68.99 & 61.36 & 62.91 & 61.73 \\
SimCSE-RoBERTa$_{\textrm{base}}$\(^\dagger\) & 70.16 &  81.77 & 73.24 & 81.36 & 80.65 & 80.22 & 68.56 & 76.57 \\
PromptRoBERTa$_{\textrm{base}}$ & \textbf{73.94$_{\pm 0.90}$} & \textbf{84.74$_{\pm 0.36}$} & \textbf{77.28$_{\pm 0.41}$} & \textbf{84.99$_{\pm 0.25}$} & \textbf{81.74$_{\pm 0.29}$} & \textbf{81.88$_{\pm 0.37}$} & \textbf{69.50$_{\pm 0.57}$} & \textbf{79.15$_{\pm 0.25} $}\\
\hline
\hline

\multicolumn{9}{c}{\textit{Supervised models}} \\
\hline
InferSent-GloVe$^\S$ & 52.86 & 66.75 & 62.15 & 72.77 & 66.87 & 68.03 & 65.65 & 65.01 \\
SBERT$_{\textrm{base}}$ $^\S$& 70.97 & 76.53 & 73.19 & 79.09 & 74.30 & 77.03 & 72.91 & 74.89 \\
SBERT$_{\textrm{base}}$-flow\(^\dagger\) & 69.78 & 77.27 & 74.35 & 82.01 & 77.46 & 79.12 & 76.21 & 76.60 \\
SBERT$_{\textrm{base}}$-whitening\(^\dagger\) & 69.65 & 77.57 & 74.66 & 82.27 & 78.39 & 79.52 & 76.91 & 77.00 \\
ConSERT$_{\textrm{base}}$\(^\ddagger \) & 74.07 & 83.93 & 77.05 & 83.66 & 78.76 & 81.36 & 76.77 & 79.37\\
SimCSE-BERT$_{\textrm{base}}$\(^\dagger\) & 75.30 & 84.67 & 80.19 & 85.40 & 80.82 & 84.25 & 80.39 & 81.57 \\
PromptBERT$_{\textrm{base}}$ & \textbf{75.48} &  \textbf{85.59} & \textbf{80.57} & \textbf{85.99} & \textbf{81.08} &    \textbf{84.56}     &      \textbf{80.52}      & \textbf{81.97} \\
\hline
SRoBERTa$_{\textrm{base}}$ $^\S$ & 71.54 & 72.49 & 70.80 & 78.74 & 73.69 & 77.77 & 74.46 & 74.21 \\
SRoBERTa$_{\textrm{base}}$-whitening\(^\dagger\) & 70.46 & 77.07 & 74.46 & 81.64 & 76.43 & 79.49 & 76.65 & 76.60 \\
SimCSE-RoBERTa$_{\textrm{base}}$\(^\dagger\) & 76.53 & 85.21 & 80.95 & 86.03 & 82.57 & 85.83 & \textbf{80.50} & 82.52 \\
PromptRoBERTa$_{\textrm{base}}$ &  \textbf{76.75} & \textbf{85.93} & \textbf{82.28} & \textbf{86.69} & \textbf{82.80} &    \textbf{86.14}
  &      80.04 & \textbf{82.95}  \\
\hline
\end{tabular}
\caption{ The performance comparison of our fine-tuned BERT methods on STS tasks.  For unsupervised models, we found the result of unsupervised constrastive learning is unstable, and we train our model with 10 random seeds. 
\(\dagger\): results from \cite{gao2021simcse}. \(\ddagger \): results from \cite{yan2021consert}. $\S $: results from \cite{reimers2019sentence}. $\P$: results from \cite{zhang2020unsupervised}.
} \label{tab:ftr}
\end{table*}

\subsection{Dataset} Following the past works \cite{yan2021consert, gao2021simcse,  reimers2019sentence}, we conduct our experiments on 7 common STS datasets: STS tasks 2012-2016 \cite{agirre2012semeval, agirre2013sem, agirre2014semeval, agirre2015semeval, agirre2016semeval} STS-B\cite{cer2017semeval}, SICK-R \cite{marelli2014sick}. We use the SentEval toolkit \cite{conneau2018senteval} to download all 7 datasets. The sentence pairs in each dataset are scored from 0 to 5 to indicate semantic similarity.

\subsection{Baselines}
We compare our method with both enlightening and state-of-the-art methods.
To validate the effectiveness of our method in the  non fine-tuned setting, we use the GLoVe \cite{pennington2014glove} and postprocess methods: BERT-flow \cite{li2020sentence} and BERT-whitening \cite{su2021whitening} as baselines.
For the fine-tuned setting, we compare our method with IS-BERT\cite{zhang2020unsupervised}, InferSent\cite{conneau2017supervised}, Universal Sentence Encoder\cite{cer2018universal}, SBERT\cite{reimers2019sentence} and the contrastive learning based methods: SimCSE \cite{gao2021simcse} and ConSERT \cite{yan2021consert}.

\subsection{Implementation Details}
 For the non fine-tuned setting, we report the result of BERT to validate the effectiveness of our representation method. For the fine-tuned setting, we use BERT and RoBERTa with the same unsupervised and supervised training data with \cite{gao2021simcse}. Our methods are trained with prompt based contrastive learning with template denosing. The templates used for both settings are manual searched according to Table \ref{tab:gss}. More details can be found in Appendix \ref{sec:appendix_td}.

\subsection{Non Fine-Tuned BERT Results}
To connect with the previous analysis of the poor performance of original BERT, we report our prompt based methods with non fine-tuned BERT in Table \ref{tab:ufr}. Using templates can substantially improve the results of original BERT on all datasets. Compared to pooling methods like averaging of last layer or averaging of first and last layers, our methods can improve spearman correlation by more than 10\%. Compared to the postprocess methods: BERT-flow and BERT-whitening, only using the manual template surpasses can these methods. Moreover, we can use the continuous template by OptiPrompt to help original BERT achieve much better results, which even outperforms unsupervised ConSERT in Table \ref{tab:ftr}.

\subsection{Fine-Tuned BERT Results}
The results of fine-tuned BERT are shown in Table \ref{tab:ftr}. Following previous works \cite{reimers2019sentence},  we run unsupervised and supervised methods respectively. Although the current contrastive learning based methods \cite{gao2021simcse,yan2021consert} achieved significant improvement compared to the previous methods, our method still outperforms them.
Prompt based contrastive learning objective significantly shortens the gap between the unsupervised and supervised methods. It also proves our method can leverage the knowledge of unlabeled data with different templates as positive pairs.
Moreover, we report the unsupervised performance with 10 random seeds to achieve more accurate results. In Discussion, we also report the result of SimCSE with 10 random seeds. Compared to SimCSE, our method shows more stable results than it.

\begin{table*}[h!]
\centering
\renewcommand\arraystretch{1.2}
\setlength{\tabcolsep}{2pt}
\begin{tabular}{lcc}
\hline
Sentence & Top-5 tokens & Top-5 tokens after template denoising \\
\hline
i am sad. & sad,sadness,happy,love,happiness & sad,sadness,crying,grief,tears\\
i am not happy. & happy,happiness,sad,love,nothing & sad,happy,unhappy,upset,angry \\
the man is playing the guitar. & guitar,song,music,guitarist,bass & guitar,guitarist,guitars,playing,guitarists\\
the man is playing the piano. & piano,music,no,yes,bass & piano,pianist,pianos,playing,guitar\\
\hline
\end{tabular}

\caption{ The top-5 tokens predicted by manual template with original BERT.
} \label{tab:etd}
\end{table*}
\subsection{Effectiveness of Prompt Based Contrastive Learning with Template Denoising}
We report the results of different unsupervised training objectives in prompt based BERT. We use the following training objectives: 1) the same template, which uses inner dropout noise as data augmentation \cite{gao2021simcse} 2) the different templates as positive pairs 3) the different templates with template denoising (our default method).
Moreover, we use the same template and setting to predict and only change the way to generate positive pairs in the training stage. All results are from 10 random runs.
The result is shown in Table \ref{tab:dot}. We observe our method can achieve the best and most stable results among three training objectives.

\begin{table}[h!]
\renewcommand\arraystretch{1.2}
\centering
\small
\setlength{\tabcolsep}{2pt}
\begin{tabular}{lcc}
\hline
    & BERT$_\textrm{base}$ & RoBERTa$_\textrm{base}$ \\
\hline
same template (dropout) & 78.16$_{\pm 0.17}$ & 78.16$_{\pm 0.44}$ \\
different templates & 78.19$_{\pm 0.29 }$ & 78.17$_{\pm 0.44}$\\
different templates with denoising & 78.54$_{\pm 0.15}$ & 79.15$_{\pm 0.25}$\\
\hline
\end{tabular}
\caption{Comparison of different unsupervised training objectives.
} \label{tab:dot}
\end{table}
\vspace{-13pt}

\section{Discussion}

\subsection{Template Denoising}
We find the template denoising efficiently removes the bias from templates and improves the quality of top-k tokens predicted by MLM head in original BERT.  As Table \ref{tab:etd} shows, we predict some sentences' top-5 tokens in the [MASK] tokens. We find the template denoising removes the unrelated tokens like ``nothing,no,yes'' and helps the model predict more related tokens. To quantify this, we also represent the sentence from the Eq. \ref{eq:anisotropy2} by using the weighted average of top-200 tokens as the sentence embeddings. The results are shown in Table \ref{tab:dsts}. The template denoising significantly improves the quality of tokens predicted by MLM head. However, it can't improve the performance for our default represent method in the Eq. \ref{eq:anisotropy1} ([MASK] token in Table \ref{tab:dsts}). In this work, we only use the template denoising in our contrastive training objective, which helps us eliminate different template biases.

\begin{table}[h!]
\renewcommand\arraystretch{1.3}
\centering
\small
\setlength{\tabcolsep}{5pt}
\begin{tabular}{lcc}
\hline
                    & no denoising & denoising\\
\hline
avg. Top-200 tokens & 56.19        & 60.39 \\
{}[MASK] token          & 67.85       & 67.43 \\
\hline
\end{tabular}
\caption{Influence of template denoising in sentence embeddings.
} \label{tab:dsts}
\end{table}
\vspace{-10pt}

\subsection{Stability in Unsupervised Contrastive Learning}

To prove the unstable results in unsupervised contrastive learning in sentence embeddings, we also reproduce the result of unsupervised SimCSE-BERT$_\textrm{base}$ with 10 random seeds in Table \ref{tab:unstable}. Our results are more stable than SimCSE. The difference between the best and worst results can be up to 3.14\% in SimCSE. However, the gap in our method is only 0.53.

\begin{table}[h!]
\renewcommand\arraystretch{1.3}
\centering
\small
\begin{tabular}{lccc}
\hline
    & Mean & Max & Min \\
\hline
SimCSE-BERT$_\textrm{base}$ &  75.42$_{\pm 0.86}$ & 76.64 & 73.50 \\
PromptBERT$_\textrm{base}$ &  78.54$_{\pm 0.15}$ & 78.86 & 78.33 \\
\hline
\end{tabular}
\caption{Results in unsupervised contrastive learning.
} \label{tab:unstable}
\end{table}
\vspace{-10pt}

\section{Conclusion}
In this paper, we analyzed the poor performance of original BERT for sentence embeddings, and find original BERT is underestimated in sentence embeddings due to inappropriate sentence representation methods.
These methods suffer from static token embedding bias and do not effectively use the original BERT layer.
To better leverage BERT in sentence embeddings, we propose a prompt-based sentence embedding method, which helps original BERT achieve impressive performance in sentence embeddings.
To further improve our method in fine-tuning, we proposed a contrastive learning method based on template denoising.
Our extensive experiments demonstrate the efficiency of our method on STS tasks and transfer tasks.

\section{Limitation}
While our methods achieve reasonable performance on both unsupervised and supervised settings,
the templates used are still manually generated. Although we have tried automatic templates generated by T5, these templates still underperform manual templates. Furthermore, we also show the performance with continuous templates, which verify the efficiency of prompts in sentence embeddings.
We expect that a carefully designed automatic template-generated mechanism can lead to higher improvement.
We leave it in the future.

\section{Acknowledgments}
The research work is supported by the National Key Research and Development Program of China under Grant No. 2021ZD0113602, the National Natural Science Foundation of China under Grant Nos. 62276015, 62176014, the Fundamental Research Funds for the Central Universities.

\bibliography{ref}
\bibliographystyle{acl_natbib}

\appendix
\section{Static Token Embeddings Biases}\label{sec:appendix_sta_token}
\subsection{Eliminating Biases by Removing Tokens}
\begin{figure*}[h]
\centering    

\subfigure[Frequency bias in static token embeddings of untying weights pre-trained model.]
{
	\includegraphics[width=0.66\columnwidth]{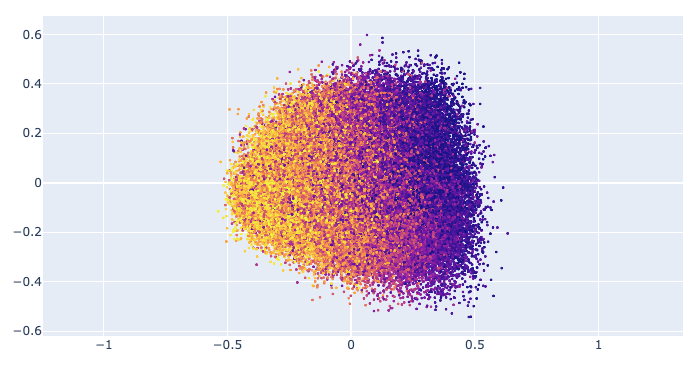}
}
\subfigure[Frequency bias in MLM head of untying weights pre-trained model.]
{
	\includegraphics[width=0.66\columnwidth]{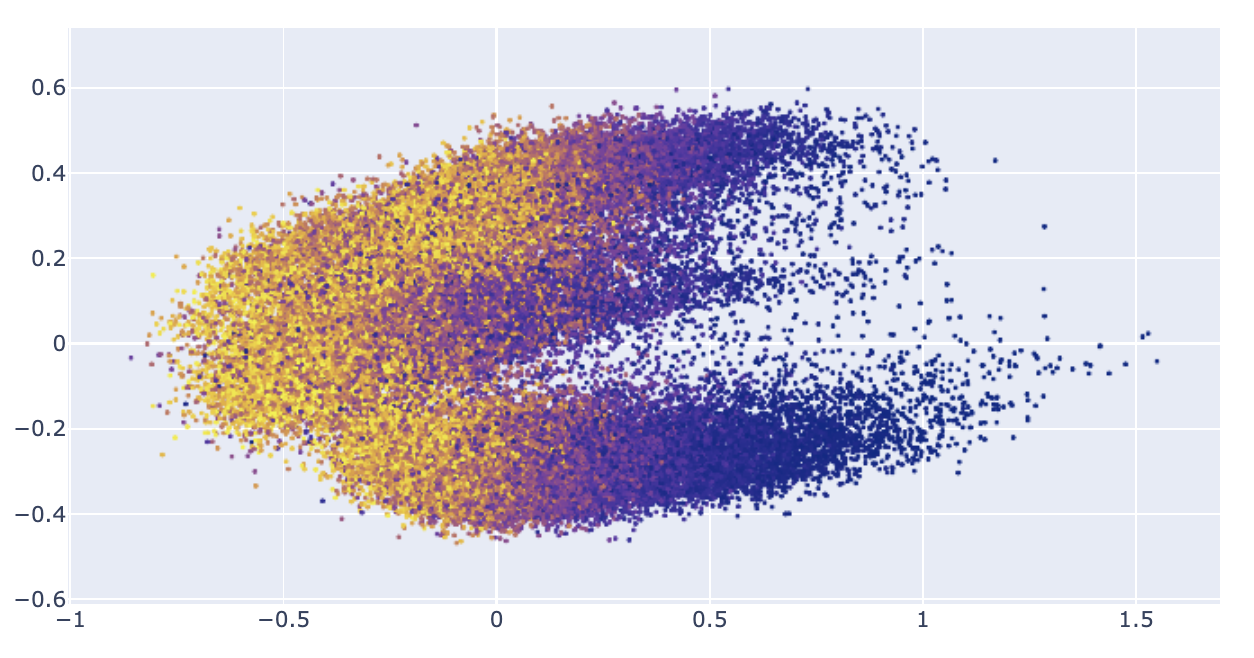}
}
\subfigure[Frequency bias in tying weights pre-trained model.]
{
	\includegraphics[width=0.66\columnwidth]{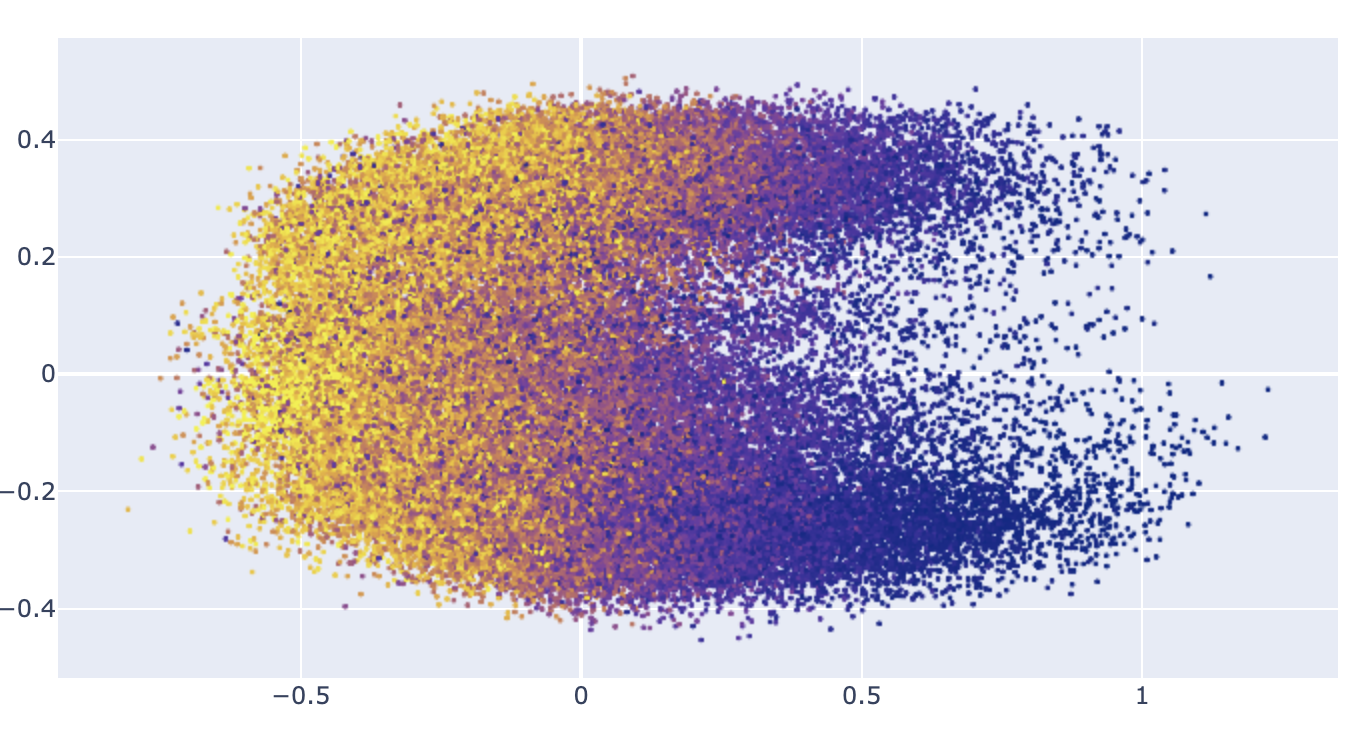}
}

\subfigure[Subword and Case biases in static token embeddings of untying weights pre-trained model.]
{
	\includegraphics[width=0.66\columnwidth]{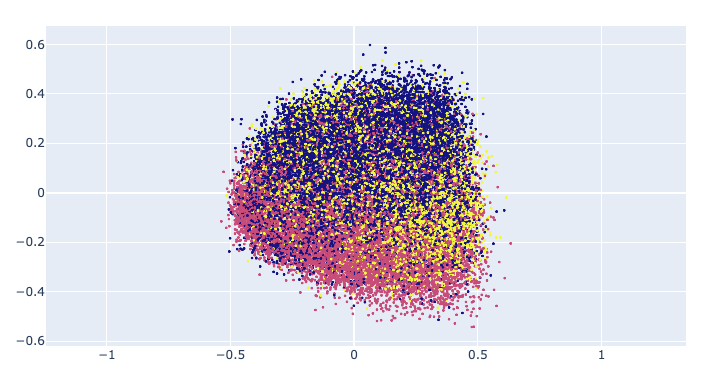}   
}
\subfigure[Subword and Case biases in MLM head of untying weights pre-trained model.]
{
	\includegraphics[width=0.66\columnwidth]{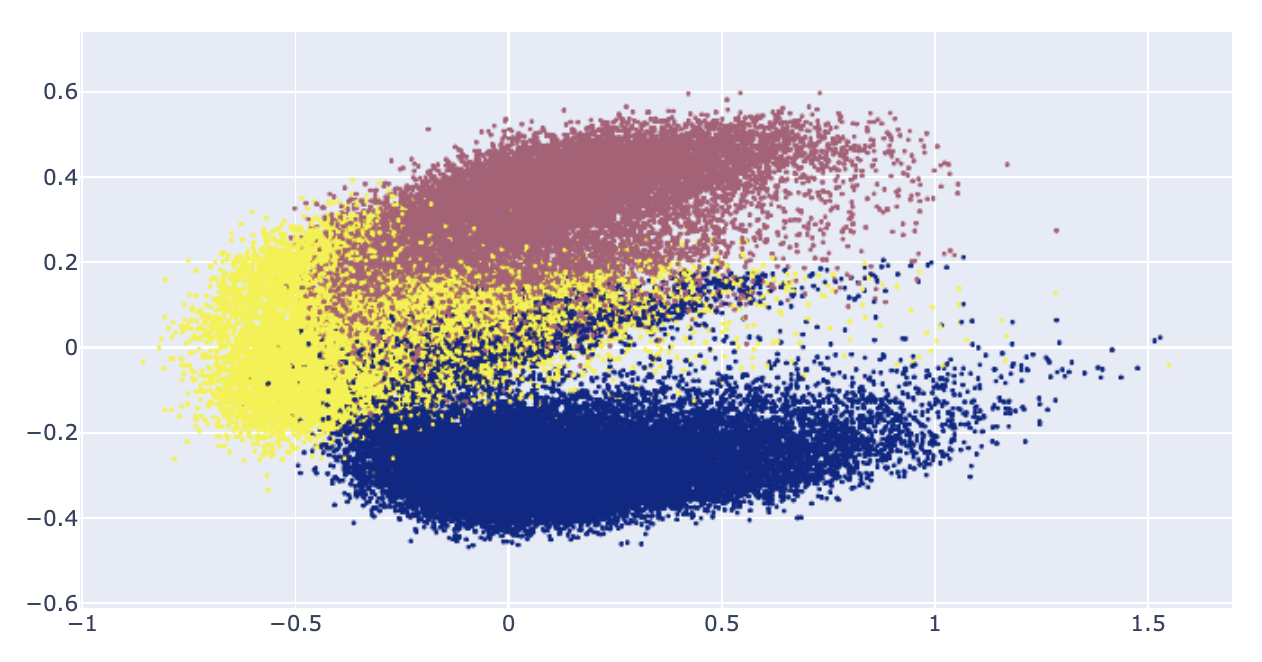}   
}
\subfigure[Subword and Case biases in tying weights pre-trained model.] 
{
	\includegraphics[width=0.66\columnwidth]{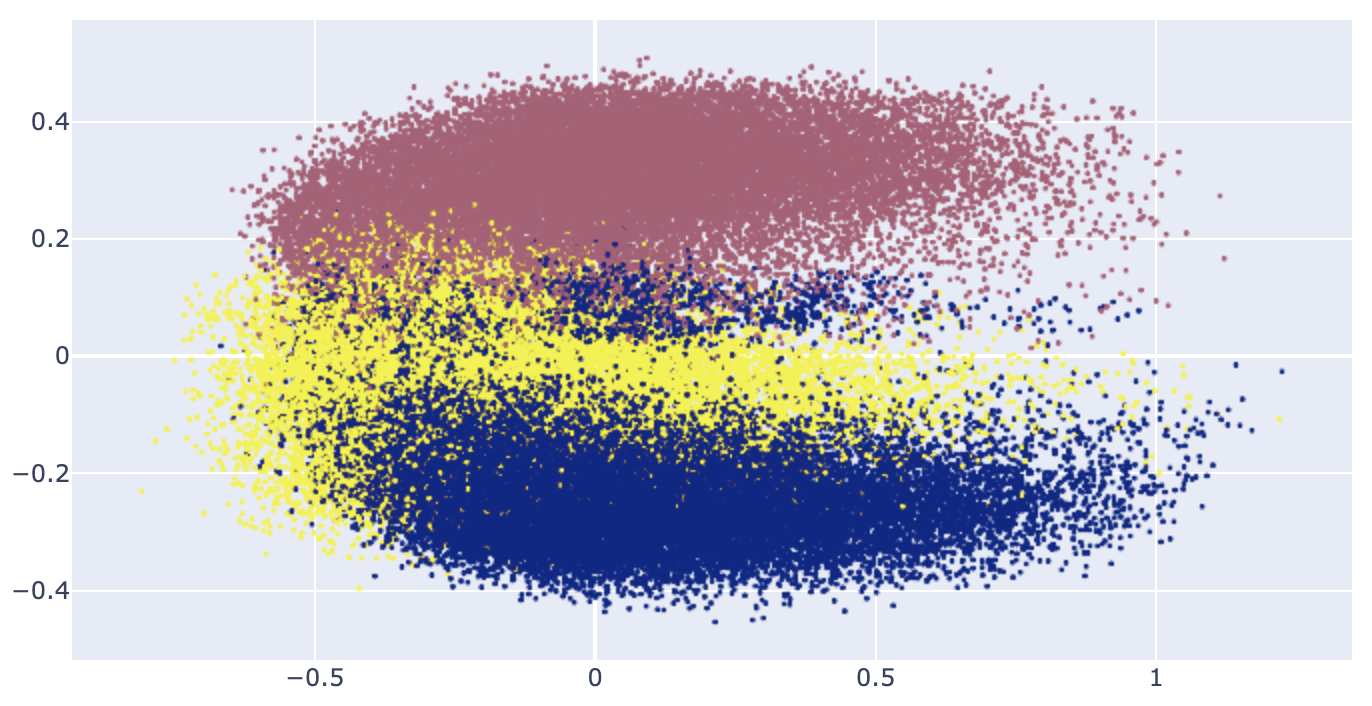}   
}

\caption{2D visualization of static token embeddings in untying and tying weights pre-trained model. For frequency bias, the darker the color, the higher the token frequency. For subword and case bias, yellow represents subword and red represents the token contains capital letters.
} 
\label{fig:uw2dv}
\end{figure*}

We reported the detailed implementation of eliminating static token embeddings biases by deleting tokens on \texttt{bert-base-uncased}, \hytt{bert-base-cased} and \texttt{roberta-base}. For Freq. tokens, we follow the settings in \cite{yan2021consert} and remove the top 36 frequent tokens. The removed Freq. tokens are shown in Table~\ref{tab:freq_tokens}. For Sub. tokens, we directly remove all subword tokens (yellow tokens in Figure \ref{fig:uw2dv}). For Case. tokens, only SICK\cite{marelli2014sick} has sentences with upper and lower case, and we lowercase these sentences to remove the uppercased tokens (red tokens in Figure \ref{fig:uw2dv}). For Pun., we remove the tokens, which contain only punctuations.

\begin{table}[h!]
\centering
\small
\setlength{\tabcolsep}{2pt}
\begin{tabular}{cl}
\hline
                                        & Removed Top frequency Tokens\\
\hline
 \hytt{bert-base-uncased}             & . a the in , is to of and ' on\\
   \multirow{2}*{and}                   & - s with for " at \#\#s woman are\\
                                        & two that you dog said playing\\
  \hytt{bert-base-cased}              &  an as was from : by white\\
\hline
   \multirow{5}*{\hytt{roberta-base}} & Ġ. Ġa Ġthe Ġin a Ġ,  Ġis Ġto Ġof\\
                                        & Ġon Ġ' s . the Ġman - Ġwith Ġfor\\
                                        & Ġwoman Ġare Ġ" Ġthat Ġit Ġdog\\
                                        & Ġplaying Ġwas Ġas Ġfrom Ġ: Ġyou\\
                                        & i Ġby\\
  \hline
\end{tabular}
\caption{Removed top 36 frequent tokens in \hytt{bert-base-cased}, \hytt{bert-base-uncased} and \hytt{roberta-base}.
}\label{tab:freq_tokens}
\end{table}

\subsection{Eliminating Biases by Pre-training }
According to \cite{gao2019representation}, we find the most of biases in static token embeddings are gradient from the MLM classification head weight, which transform the last hidden vector of [MASK] to the probability of all tokens. The tying weight between the static token embeddings and MLM classification head causes static token embeddings to suffer from bias problems.

We have pre-trained two BERT-like models with the MLM pre-training objective. The only difference between the two pre-trained models is tying and untying the weight between static token embeddings and MLM classification head. We have pre-trained these two models on 125k steps with 2k batch sizes.

As shown in Figure~\ref{fig:uw2dv}, we have shown the static token embeddings of the untying model, MLM head weight of untying model and static token embeddings (MLM head weight) of the tying model. The distribution of the tying model and the head weight of the untying model is same with \hytt{bert-base-cased} in Figure~\ref{fig:2dv}, which severely suffers from the embedding biases. However, the distribution of the token embeddings in the untying weights model is less influenced by these biased. We also report the average spearman correlation of three embedding on STS tasks in Table~\ref{tab:ttreuslt}. Static token embeddings of the untying model achieves the best correlation among the three embeddings.

\begin{table}[h]
\centering
\small
\setlength{\tabcolsep}{2pt}
\begin{tabular}{lc}
\hline
                                        &  Avg.\\
\hline
MLM head of untying model               & 43.33 \\
Static token embeddings of untying model & 49.41 \\
Static token embeddings of tying model   & 45.68 \\
\hline
\end{tabular}
\caption{ The avg. spearman correlation of three embeddings.
}\label{tab:ttreuslt}
\end{table}

\section{Training Details}\label{sec:appendix_td}

\begin{table}[h!]
\centering
\small
\setlength{\tabcolsep}{2pt}
\begin{tabular}{ccccc}
\hline
Model & Template\\
\hline
\multirow{2}*{BERT} & This sentence of ``[X]'' means [MASK] . \\
& This sentence : ``[X]'' means [MASK] . \\
\hline
\multirow{2}*{RoBERTa}    & This sentence : `[X]' means [MASK] . \\
& The sentence : `[X]' means [MASK] . \\
  \hline
\end{tabular}
\caption{Templates for our method in fine-tuned setting}
\label{tab:templates}
\end{table}

\begin{table}[h!]
\centering
\small
\setlength{\tabcolsep}{2pt}
\begin{tabular}{ccccc}
\hline
& \multicolumn{2}{c}{\textit{Unsupervised}} & \multicolumn{2}{c}{\textit{Supervised}} \\
                & BERT & RoBERTa & BERT & RoBERTa\\
\hline
Batch size    & 256  & 256  & 512  & 512 \\
Learning rate & 1e-5 & 1e-5 & 5e-5 & 5e-5 \\
Epoch         & 1    & 1    & 3    & 3 \\
Vaild steps   & 125  & 125  & 125  & 125\\
  \hline
\end{tabular}
\caption{Hyperparameters for our method in fine-tuned setting}
\label{tab:hyper}
\end{table}

\begin{table*}[t]
\renewcommand\arraystretch{1.2}
\centering
\small
\setlength{\tabcolsep}{5pt}
\begin{tabular}{lcccccccc}
\hline
Method                     & MR             & CR             & SUBJ           & MPQA           & SST-2          & TREC           & MRPC           & Avg.\\
\hline
\hline
\multicolumn{9}{c}{\textit{Unsupervised models}} \\
\hline
Avg. BERT embedding        & 78.66          & 86.25          & 94.37          & 88.66          & 84.40          & 92.80          & 69.54          & 84.94 \\
BERT-[CLS] embedding       & 78.68          & 84.85          & 94.21          & 88.23          & 84.13          & 91.40          & 71.13          & 84.66 \\
IS-BERT                    & 81.09          & \textbf{87.18} & \textbf{94.96} & 88.75          & \textbf{85.96} & 88.64          & 74.24          & 85.83 \\
SimCSE-BERT                & \textbf{81.18} & 86.46          & 94.45          & 88.88          & 85.50          & \textbf{89.80} & 74.43          & \textbf{85.81} \\
PromptBERT                 & 80.74          & 85.49          & 93.65          & \textbf{89.32} & 84.95          & 88.20          & \textbf{76.06} & 85.49\\
\hline
SimCSE-RoBERTa             & 81.04          & 87.74          & \textbf{93.28} & 86.94          & 86.60          & 84.60          & 73.68          & 84.84 \\
PromptRoBERTa              & \textbf{83.82} & \textbf{88.72} & 93.19          & \textbf{90.36} & \textbf{88.08} & \textbf{90.60} & \textbf{76.75} & \textbf{87.36} \\
\hline
\hline
\multicolumn{9}{c}{\textit{Supervised models}} \\
\hline
InferSent-GloVe            & 81.57          & 86.54          & 92.50          & 90.38          & 84.18          & 88.20          & 75.77          & 85.59\\
Universal Sentence Encoder & 80.09          & 85.19          & 93.98          & 86.70          & 86.38          & 93.20          & 70.14          &85.10\\
SBERT                      & \textbf{83.64} & \textbf{89.43} & 94.39          & 89.86          & \textbf{88.96} & \textbf{89.60} & 76.00          & \textbf{87.41} \\
SimCSE-BERT                & 82.69          & 89.25          & \textbf{94.81} & 89.59          & 87.31          & 88.40          & 73.51          & 86.51\\
PromptBERT                 & 83.14          & 89.38          & 94.49          & \textbf{89.93} & 87.37          & 87.40          & \textbf{76.58} & 86.90 \\
\hline
SRoBERTa                   & 84.91          & 90.83          & 92.56          & 88.75          & 90.50          & 88.60          & \textbf{78.14} & 87.76\\
SimCSE-RoBERTa             & 84.92          & \textbf{92.00} & 94.11          & 89.82          & 91.27          & 88.80          & 75.65          & 88.08 \\
PromptRoBERTa              & \textbf{85.74} & 91.47          & \textbf{94.81} & \textbf{90.93} & \textbf{92.53} & \textbf{90.40} & 77.10          & \textbf{89.00} \\
\hline
\hline
\end{tabular}
\caption{ Transfer task results of different sentence embedding models.
} \label{tab:ttr}
\end{table*}

For the non fine-tuned setting, the manual template we used is \textit{This sentence : ``[X]'' means [MASK] .}. For OptPrompt, we first initialize the template embeddings with the manual template and then train these template embeddings by freezing BERT with the unsupervised training task followed by \cite{gao2021simcse}, and the batch size, learning-rate, epoch and valid steps are 256, 3e-5, 5 and 1000.

For the fine-tuned setting,
all training data is same with \cite{gao2021simcse}. The max sentence sequence length is set to 32.
For templates, we only use the manual templates, which are manually searched according to STS-B dev in unfine-tuned models. The templates is shown in Table \ref{tab:templates}.
For unsupervised method, we use two different templates for unsupervised training with template denosing according to our prompt based training objective. In predicting, we directly use the one template without template denoising.
For supervised method, we use template denoising with same template for contrastive learning, because we already have supervised negative samples.  
We also report other training details in Table \ref{tab:hyper}.

\section{Transfer Tasks}\label{sec:appendix_tt}
We also evaluate our models on the following transfer tasks: MR, CR, SUBJ, MPQA, SST-2, TREC and MRPC. We follow the default configurations in SentEval\footnote{https://github.com/facebookresearch/SentEval}. The results are shown in Table~\ref{tab:ttr}.
Comparing to SimCSE, our RoBERTa based method can improve 2.52 and 0.92 on unsupervised and supervised models respectively. 

\end{document}